\documentclass[letterpaper, 10 pt, conference]{ieeeconf}  

\IEEEoverridecommandlockouts

\usepackage{cite}
\usepackage{amsmath,amssymb,amsfonts}
\usepackage{algorithmic}
\usepackage{graphicx}
\usepackage{textcomp}
\usepackage{xcolor}

\usepackage{caption}
\usepackage{subcaption}
\usepackage{hyperref}
\usepackage{algorithm}

\usepackage{array}

\usepackage{listings}

\usepackage{booktabs}
\usepackage{multirow}
\usepackage{siunitx}

\def\BibTeX{{\rm B\kern-.05em{\sc i\kern-.025em b}\kern-.08em
    T\kern-.1667em\lower.7ex\hbox{E}\kern-.125emX}}

\title{OceanChat: Piloting Autonomous Underwater Vehicles in Natural Language}

\author{Ruochu Yang$^{1}$, Mengxue Hou$^{2}$, Junkai Wang$^{1}$, Fumin Zhang$^{1}$
\thanks{This research work is supported by ONR grants N00014-19-1-2556 and N00014-19-1-2266;  AFOSR grant FA9550-19-1-0283; NSF grants GCR-1934836,  CNS-2016582 and ITE-2137798; and NOAA grant NA16NOS0120028.}
\thanks{$^{1}$Ruochu Yang, Junkai Wang, and Fumin Zhang are with School of Electrical and Computer Engineering, Georgia Institute of Technology, Atlanta, USA. $^{2}$Mengxue Hou is with School of Electrical Engineering, University of Notre Dame, Notre Dame, USA.}
}

\begin{document}

\maketitle
\thispagestyle{empty}
\pagestyle{empty}

\begin{abstract}
In the trending research of fusing Large Language Models (LLMs) and robotics, we aim to pave the way for innovative development of AI systems that can enable Autonomous Underwater Vehicles (AUVs) to seamlessly interact with humans in an intuitive manner. We propose OceanChat, a system that leverages a closed-loop LLM-guided task and motion planning framework to tackle AUV missions in the wild. LLMs translate an abstract human command into a high-level goal, while a task planner further grounds the goal into a task sequence with logical constraints. To assist the AUV with understanding the task sequence, we utilize a motion planner to incorporate real-time Lagrangian data streams received by the AUV, thus mapping the task sequence into an executable motion plan. Considering the highly dynamic and partially known nature of the underwater environment, an event-triggered replanning scheme is developed to enhance the system's robustness towards uncertainty. We also build a simulation platform HoloEco that generates photo-realistic simulation for a wide range of AUV applications. Experimental evaluation verifies that the proposed system can achieve improved performance in terms of both success rate and computation time. Project website: \url{https://sites.google.com/view/oceanchat}    
\end{abstract}

\section{Introduction}
\label{intro}

   \begin{figure*}
        \centerline{\includegraphics[width=\textwidth]{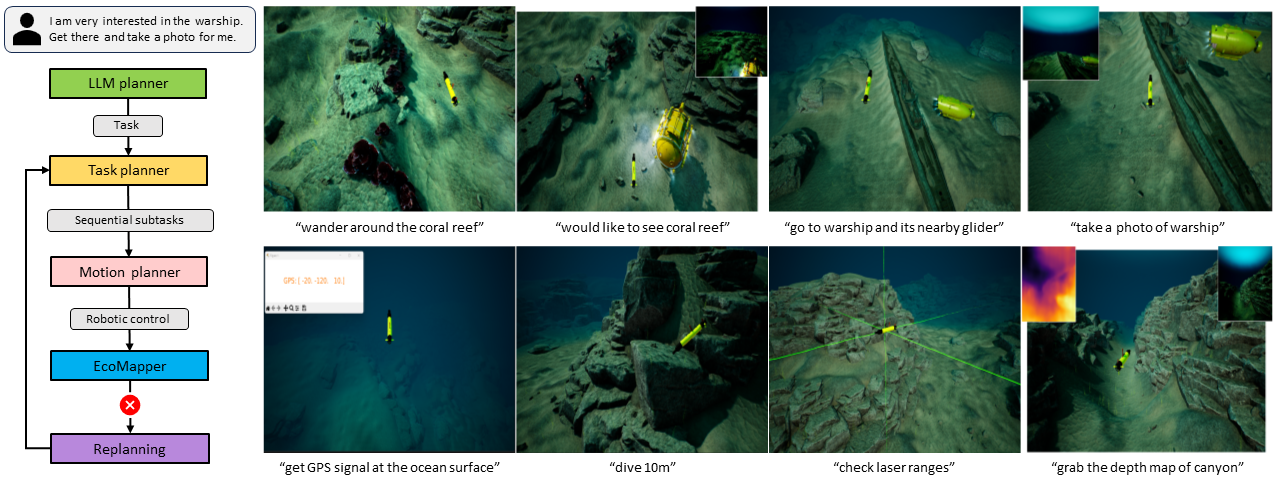}}
        \caption{Our proposed system \textbf{OceanChat} can decompose \textbf{natural language commands} into a task sequence of \textbf{controlling EcoMapper} in the HoloEco simulation platform. We establish a three-level framework of LLM-guided task and motion planner. The event-triggered replanning module is designed to withstand unpredictable error during execution.}
        \label{whole system}
    \end{figure*}

AUVs have been widely used in ocean engineering for a range of applications, including algal bloom monitoring, hurricane prediction, underwater acoustics, and ocean observation systems \cite{doi:10.1080/00207170701222947, 4476150, nicholson2008present}. However, piloting AUVs during real-world missions is usually laborious with high demand of mechanical manuals, mission configuration files, and terminal commands. From the perspective of a operator, it would be ease to simplify AUV piloting processes by abstracting technical complexities with natural language. Moreover, it is fairly intriguing for us to yearn the possibility of making AUVs handful for everyone. Rather than requiring a specialized engineer to control the AUV, our blueprint is to have a non-technical user on the loop, i.e., deploying underwater missions through natural language. The emergence of LLMs offers a promising avenue to achieve this vision, as they can learn to project real-world concepts into the language space. While LLMs are believed to dig out open-world knowledge in the text form, it remains a critical aspect of using such knowledge to enable robots to physically act in the real world \cite{tellex2020robots, bollini2013interpreting, tellex2011understanding}. The question then arises: how can we ground abstract language instructions into  AUVs' physical actions? For example, given an AUV pilot's command "go through the canyon", how can LLMs  trigger basic AUV controllers and sensors, such as moving forward and taking photos, to accomplish the overarching goal?

We propose a system OceanChat, which is able to pilot AUVs in natural language as shown in Fig.~\ref{whole system}.  The proposed system leverages a closed-loop LLM-guided task and motion planning framework to perform challenging AUV missions and replan in case of execution failure. The main contributions of this paper are summarized as follows.

\begin{itemize}
    \item  We develop OceanChat, a simulator and planner for using natural language to control a calibrated AUV agent EcoMapper.
    \item  We establish a closed-loop LLM-guided task and motion planning framework to endow OceanChat with executable robotic control.
\end{itemize}

This paper is organized as follows. Section \ref{related works} outlines related works. Section \ref{HoloEco platform} presents the HoloEco simulation platform along with the refined AUV model EcoMapper. Section \ref{method} illustrates the OceanChat system featuring the closed-loop LLM-task-motion planning framework. Section \ref{experimental evalution} evaluates OceanChat in terms of achieving underwater missions given human commands. Section \ref{conclusion} provides conclusion and future works.

\section{Related Works}
\label{related works}

LLMs have exhibited considerable capabilities of deciphering natural language within the context of real-world scenarios, including language understanding, sentiment analysis, text completion, etc. Recently, harnessing LLMs for robotic applications has emerged as a rapidly evolving field of research. One notable significance of connecting LLMs with robotics is to allow robotic agents to interact with the world and  humans in a natural way. A large body of works aim to integrate LLMs into a planning and reasoning pipeline for robotic execution. One commonly affordable approach relies on  prompting strategies for LLMs to derive a sequential plan aimed at achieving a user prompt \cite{huang2022language, singh:progprompt}. \cite{vemprala2023chatgpt} maps  LLM-guided steps to pre-deﬁned robot skills by listing a collection of high-level functions in the prompt. \cite{ahn2022i} uses LLMs to score a candidate skill with the highest probability of completing the overall instruction. \cite{liang2023code, silver2023generalized} utilizes LLMs to create code for undefined functions, thus generalizing to unseen instructions in different robotic scenarios. Some works also encourage LLMs to demonstrate their chain of thought \cite{wei2022chain} or supplement plan explanations in a structured JSON format \cite{wake2023chatgpt}, which enables LLMs to perform step-by-step reasoning and allows users to rectify infeasible LLM responses. In the robotics realm, the conventional open-loop system pertains to executing tasks without actively sensing the environment or responding to possible failure. While this approach offers advantages of simplicity and speed, it is accompanied by several limitations including the absence of error correction mechanisms and vulnerability to disturbances \cite{garrett2020online}. Closed-loop planning serves as a suitable solution of these challenges, spanning from directly re-prompting LLMs with corrective instructions \cite{raman2022planning, sharma2022correcting}, involving real-time environmental feedback \cite{huang2022inner}, to integrating multiple modalities such as vision and touch \cite{huang2023voxposer, openai2023gpt4, driess2023palme}. However, a significant barrier still lies in building an intrinsic connection between the LLM model and the AUV. To bridge this gap between open-ended human commands and executable robotic actions, we resort to a well-established planning track in the robotics discipline.

Task and Motion Planning (TAMP) is a vastly investigated problem in the robotics community \cite{ghallab_nau_traverso_2016, cambon2009hybrid, garrett2021integrated}. Classical works address TAMP in  deterministic and fully observable space, branching out into topics like pick-place planning \cite{lozano1989task, gualtieri2021robotic}, manipulation planning \cite{simeon2004manipulation, wong2013manipulation}, navigation \cite{stilman2005navigation, thrun2000probabilistic, van2011lqg}, and rearrangement planning \cite{king2016rearrangement, ding2023task}. \cite{garrett2018ffrob, srivastava2014combined} efficiently  incorporate geometric or kinematic constraints with the heuristic plan search. \cite{kaelbling2011hierarchical, fritz2009computing} present a regression-based framework by generating goal regression and pre-images in a reversible logical chain. The Planning Domain Definition Language (PDDL) \cite{fox2003pddl2, McDermott1998PDDLthePD} standardizes formulations of AI planning, thus providing a universal interface of TAMP planners regardless of domains. Furthermore, it is a fundamental extension to consider inevitable uncertainty in real-world  planning, which derives from stochasticity or partial observability of object states \cite{dogar2012planning, lozano1984automatic}. Generally, belief-space planning needs to address two types of uncertainty: future-state uncertainty \cite{alterovitz2007stochastic, levihn2013hierarchical}, and current-state uncertainty \cite{smallwood1973optimal, kurniawati2008sarsop}. One distinguished  approach of solving TAMP in belief space is to temporally decompose long-horizon problems into a sequence of short horizons in an interleaved manner \cite{kaelbling2013integrated, curtis2022long}. On basis of this approach, \cite{hou2023interleaved} develops a bi-level algorithm leveraging the Depth First Search to achieve lower computation cost and guaranteed optimality.

It should be noted that conducting empirical algorithms with real-world AUVs is impractical, as this could  cause AUVs to drift to unexpected areas or even totally abort. Due to huge expense of AUV field trials, the imperative for a high-fidelity underwater robotics simulator becomes evident, serving as an algorithm testing tool. Numerous underwater simulators \cite{cook2014survey, sehgal2010multi} have actively come into view, some of which equipped with various agents, sensors, and communication models \cite{gwon2017development, demarco2015computationally}. MarineSIM \cite{senarathne2010marinesim} focuses on multi-agent acoustic communications. UUV Simulator \cite{Manhaes_2016} possesses accurate dynamic models and easy set-up. UWSim \cite{6385788} is an open-source simulator with multi-agent and sonar support. HoloOcean \cite{Potokar22icra} is an mature one with full support of underwater robotics, light package dependencies, and ease of adding new assets.

\section{HoloEco Simulation Platform}
\label{HoloEco platform}

We establish an all-encompassing ocean simulation platform HoloEco upon HoloOcean \cite{Potokar22icra}, which offers a rich environment of diverse underwater activities. Included within it are a myriad of objects such as coral reefs, gliders, warships, underwater mountains and canyons, etc. The primary motivation behind developing HoloEco is to prevent AUVs from being exposed to empirical setbacks in the real-world ocean. Thus, it is indispensable for HoloEco to create a simulated environment where we can continually enhance quality of AUV solutions. We make available a suite of underwater applications in which a real-world AUV might engage, such as surveying coral reefs or navigating through an unknown canyon. These provisions contribute to the platform's realistic simulation of underwater missions. Additionally, our simulator involves a finely tuned AUV model EcoMapper \cite{wang2015dynamic} that closely aligns with its real-world counterpart. EcoMapper is meticulously crafted with efficient controllers and multimodal sensors like moving forward and grabbing depth maps. This level of detailed functionalities can present a solid claim of how AUVs can be similarly controlled in real-world missions.

\section{Method}
\label{method}

To accomplish AUV missions specified by human commands, we develop for OceanChat a three-level framework of closed-loop LLM-guided task and motion planning. Starting from a human command, the framework operates in three stages: LLMs interpreting the command, task planners sequencing tasks, and motion planners controlling EcoMapper in the ocean environment.

\subsection{Hierarchy of Closed-loop LLM-task-motion Planning}

\begin{figure*}
        \centerline{\includegraphics[width=\textwidth]{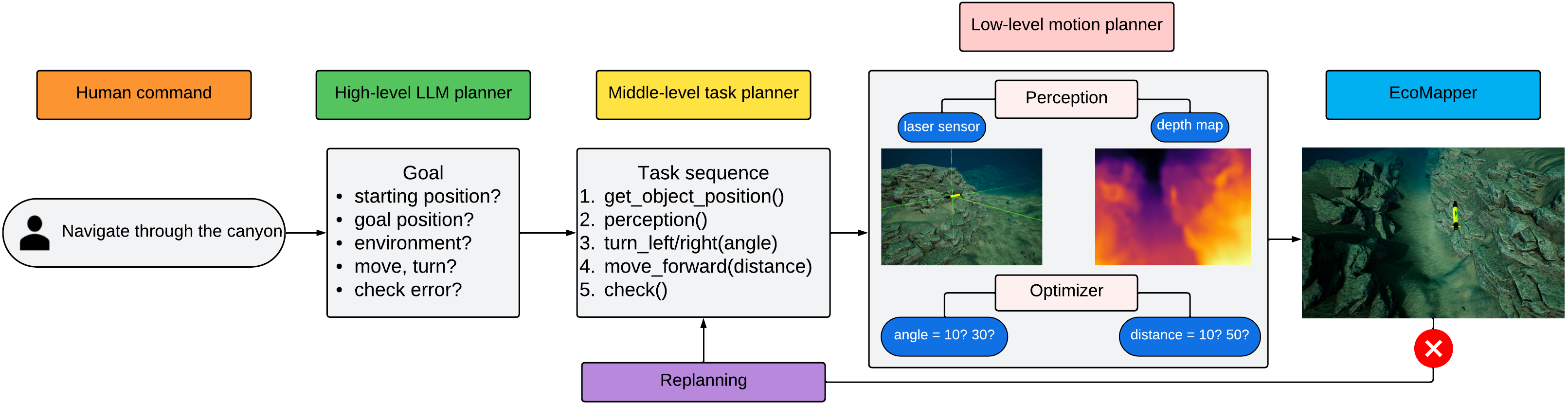}}
        \caption{Framework of closed-loop LLM-guided task and motion planning. The high-level LLM planner composes an overall goal by precisely comprehending the human command in the context of underwater scenes. The middle-level task planner generates a logical sequence of tasks to achieve the overall goal. The low-level motion planner maps the task sequence to EcoMapper robotic control by utilizing perception and optimization models. In case of execution error, the replanning module makes real-time adjustments in a closed-loop fashion.}
        \label{framework}
    \end{figure*}

Numerous gaps must be bridged all the way down from natural language commands to real robotic execution. For example, LLMs may generate an infeasible goal such as directly traversing an unknown canyon in a straight line. Task planners may propose an incorrect sequence of tasks such as moving forward a long distance before detecting the unexplored surroundings. Motion planners might compose roundabout actions that waste significant time circling inside the canyon. We demonstrate that the proposed three-level planning framework can bridge these gaps in a hierarchical manner by enhancing representations of AUV and environment dynamics level by level. The high-level LLM planner interprets the human command to offer the middle-level task planner an overall goal. The task planner then generates a feasible task sequence for the goal, which the low-level motion planner follows to enable robotic control over EcoMapper. The whole framework is shown in Fig.~\ref{framework}.

Specifically, the high-level LLM planner translates the human command into a \textit{goal} formatted as a starting state, a ending state, and a set of tasks for EcoMapper to achieve. By incorporating EcoMapper's capabilities, underwater objects, and abstraction of environmental maps, we tailor prompts to  the LLM planner so that it  can gain a symbolic understanding of the underwater scene, i.e., an abstracted representation of AUV and environment dynamics. The LLM planner can also seek clarification for ambiguous queries and outline intermediate steps by going through chain of thought. It can also achieve generalized interpretation of unseen human commands through few-shot prompting.

Although the LLM planner is filled with semantic information, it still lacks information about the physical environment. It remains unresolved whether the goal can be actually achieved and how it should be translated into a feasible task sequence (plan) for EcoMapper in the uncertain underwater environment. Given the spatial and temporal scale of AUV applications, a task sequence often extends across long planning horizons. Simultaneously, the rich action space of AUV introduces computational complexity when the task planner searches for an effective plan. Therefore, to ground the LLM-generated goal into an executable plan, we propose a middle-level task planner that refines the goal into an abstracted \textit{task sequence}. At this middle level, we represent tasks by logical representations, i.e., preconditions and effects that can be evaluated as true or false. In this way, the middle-level task planner supplements the LLM-guided goal with a task sequence based on predicate transitions learned from AUV dynamics and sensor observation. 

Finally, the low-level motion planner solves for a \textit{motion plan} to accomplish the task sequence given by the middle-level task planner. On the representation side, the motion planner supports the task planner by generating physical representations of AUV and environment dynamics from prior knowledge or real-time collected Lagrangian data streams. On the planning side, the motion planner optimizes continuous robotic control by taking physical dynamics and environmental observation into consideration, thus adding real-world constraints to the middle-level task sequence. 

Despite meticulous planning, EcoMapper may still fall into unexpected consequences owing to observation and action uncertainties in the underwater environment. Therefore, it is crucial to incorporate an event-triggered replanning module to address execution error with real-time feedback. Triggered by a logical state transition, the replanning module assesses EcoMapper's status. If the status deviates from the expected outcome, the task planner instructs the motion planner to execute corrective actions. 

\subsection{Design of Middle-level Task Planner}
\label{design of middle-level task planner}

Considering that AUVs cover hundreds of kilometers distances with limited sensors and unpredictable environmental factors, we must account for uncertainties in the real-world ocean to literally fulfill the LLM-interpreted goal. 
On one side of validity, while LLMs possess some grade of planning capability, merely relying on them as task planners will raise issues. Specifically, LLMs struggle to determine proper transitional timing for a series of specific tasks due to limited knowledge about real-world dynamics.  We refer to this pipeline as the LLM planning method. On the other side of efficiency, if we let the LLM planner directly guide the motion planner at the level of physical representations, the LLM planner will likely generate redundant and invalid attempts of combining primitive actions to navigate through the canyon. Evaluating such a tremendous number of motion trajectories is time-consuming in terms of real-time operation. We refer to this pipeline as the LLM-motion planning method. In order to achieve a good balance between validity and efficiency, we turn to establishing a middle-level task planner as a pivot to connect the LLM planner with the motion planner. The task planner can provide additional logical representations of the world to produce a feasible plan over the goal's extended horizons and pass the plan to low-level robotic controllers or sensors. This pipeline is our proposed LLM-task-motion planning method.

We formulate a generic TAMP problem $\Pi = <\Omega, \mathcal{A}, \textbf{x}_0, S_g >$  by a state space $\Omega$, an action space $\mathcal{A}$, an initial state $\textbf{x}_0 \in \Omega$ and a set of goal states $S_g \subseteq \Omega$. Let  $\textbf{x}_k \in \Omega $ and $\textbf{u}_k \in \mathcal{A}$ denote the system state and action at timestamp $k$, respectively. The goal of TAMP is to drive the  state $\textbf{x}_k$ to be inside the goal set $S_g$ under a plan of sequential actions $\pi = [\textbf{u}_0, ..., \textbf{u}_N]$. Since  there  may  be an infinite number of feasible plans, we minimize the cost function $J$ of moving along the trajectory as follows:
\begin{equation} \label{TAMP}
  \pi^* = \arg \min_{\pi} J= \sum_{k=0}^N L(\textbf{x}_k,\textbf{u}_k),
\end{equation}
where  $L$ is a one-step cost function, and the optimal plan $\pi^*$ should yield the minimum cost $J^*$. To solve the TAMP problem \ref{TAMP}, the task planner generates a task sequence to decompose a long-horizon planning problem into short horizons, and the motion planner then optimizes the specific actions of each task.

\subsubsection{HTN Task Planner}
The task planner should decide a viable sequence out of pre-defined tasks. The key aspect of designing such a task planner is to construct logical representations that include preconditions necessary to activate the tasks and effects of executing the tasks. In this work, we employ the HTN task planner to solve for the logical constraints generated by preconditions and effects of these tasks in order to achieve a target goal. In a general HTN planner, tasks are defined as higher level actions composed of primitive actions that alter Boolean-valued predicates representing the state space. Through this abstraction, the HTN planner is able to construct a task sequence that solves a complex goal without deep search.

\subsubsection{A* Motion Planner}
Within a given task sequence, the motion planner aims to find the optimal way of executing the tasks with minimum cost. Since EcoMapper mostly maintains a constant depth during operation, we introduce its dynamics as a unicycle model:
\begin{equation} \label{equation dynamics}
\begin{split}
     & \Dot{x} =  v \cos(\theta)\\
     &  \Dot{y} = v \sin(\theta)\\
     & \Dot{\theta} = \omega,
\end{split}
\end{equation}
where $x$ and $y$ are  EcoMapper's Cartesian  coordinates,  $\theta$ is  the  orientation, $v$ is the  forward velocity, and $\omega$ is the turning rate. Thus, the system state is defined as $\textbf{x} = (x, y, \theta)^T$, and the  control input is defined as $\textbf{u} = (v, \omega)^T$. We represent the dynamics model \ref{equation dynamics} in a discrete way as 
\begin{equation}
     \textbf{x}_{k+1}  = f(\textbf{x}_k, \textbf{u}_k).
\end{equation}

Let $R= \{r_1, r_2,..., r_M\} $ denote the regions that EcoMapper travels through. Given the EcoMapper position $\textbf{x}_0$ in the current region $r_i, \forall i = 1,2,...,M$, the dynamics model, and the region's environment map $V$, we can formulate the motion planning problem as an optimization problem as follows:
\begin{equation} \label{equation optimal motion planning}
\begin{split}
   & \min_{[\textbf{u}_0, ..., \textbf{u}_N]} J  = \sum_{k=0}^N L(\textbf{x}_k, \textbf{u}_k)   \\
   & \textit{s.t.  }    \textbf{x}_{k+1}  = f(\textbf{x}_k, \textbf{u}_k),   k = 0,1,..., N-1  \\
   & V, \textbf{x}_0 \in r_i, i = 1,2,...,M.
\end{split}
\end{equation} 
In this work, we leverage the A* algorithm to solve the problem \ref{equation optimal motion planning} for computing the best action.

\section{Experimental Evaluation}
\label{experimental evalution}

\begin{figure}
     \centerline{\includegraphics[width=0.48\textwidth]{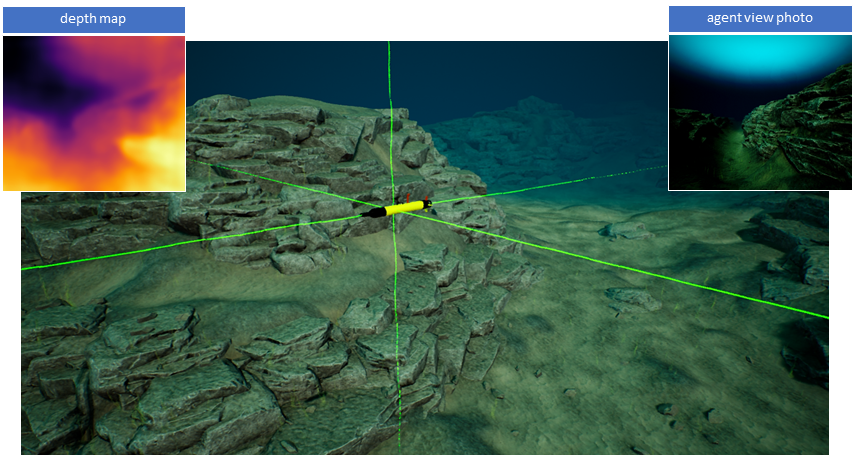}}
        \caption{Multimodal perception consisting of depth maps and laser range sensors. \textbf{Top right}: a photo in agentview of  EcoMapper. \textbf{Top left}: an inferred depth map from the photo. \textbf{Middle}: green lines emitted by the laser sensors.}
        \label{perception}
\end{figure}

\begin{figure*}
        \centerline{\includegraphics[width=\textwidth]{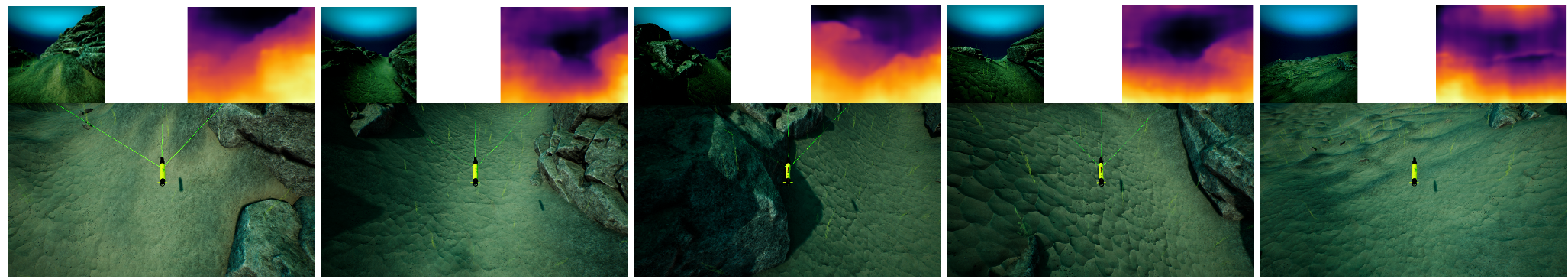}}
        \caption{\textit{Canyon Navigation}: OceanChat interprets the human command and autonomously navigates EcoMapper through the unknown canyon. The bottom figures are the viewport capture of the HoloEco scenes. The top left figures are the agentview capture of the EcoMapper. The top right figures are the depth map generated from the agentview capture. The green lines are the laser emitted by the range sensors.}
        \label{canyon navigation}
    \end{figure*}

In order to validate our system's ability of executing human commands upon AUVs, we conduct experimental evaluation in the simulated underwater world. We deliberately design a comprehensive mission of EcoMapper autonomously navigating through an unknown canyon. We know the canyon's starting and ending positions, but we lack information of the tomography inside the canyon. Due to unavailability of HD maps and GPS localization in the ocean, we develop multimodal perception consisting of depth maps and laser range sensors as shown in Fig.~\ref{perception}. Since the UDepth network \cite{yu2023udepth} offers fast depth prediction on power-limited AUVs, it is employed to generate depth maps for coarse environmental perception. For precise distance measurement, we install laser range sensors on EcoMapper in eight directions: forward, down, 30-degree left, 45-degree left, 60-degree left, 30-degree right, 45-degree right, and 60-degree right. The results of successfully navigating through the canyon are shown in Fig.~\ref{canyon navigation}. A full video is available on the \href{https://sites.google.com/view/oceanchat}{project website}. 

\begin{figure}
     \centerline{\includegraphics[width=0.48\textwidth]{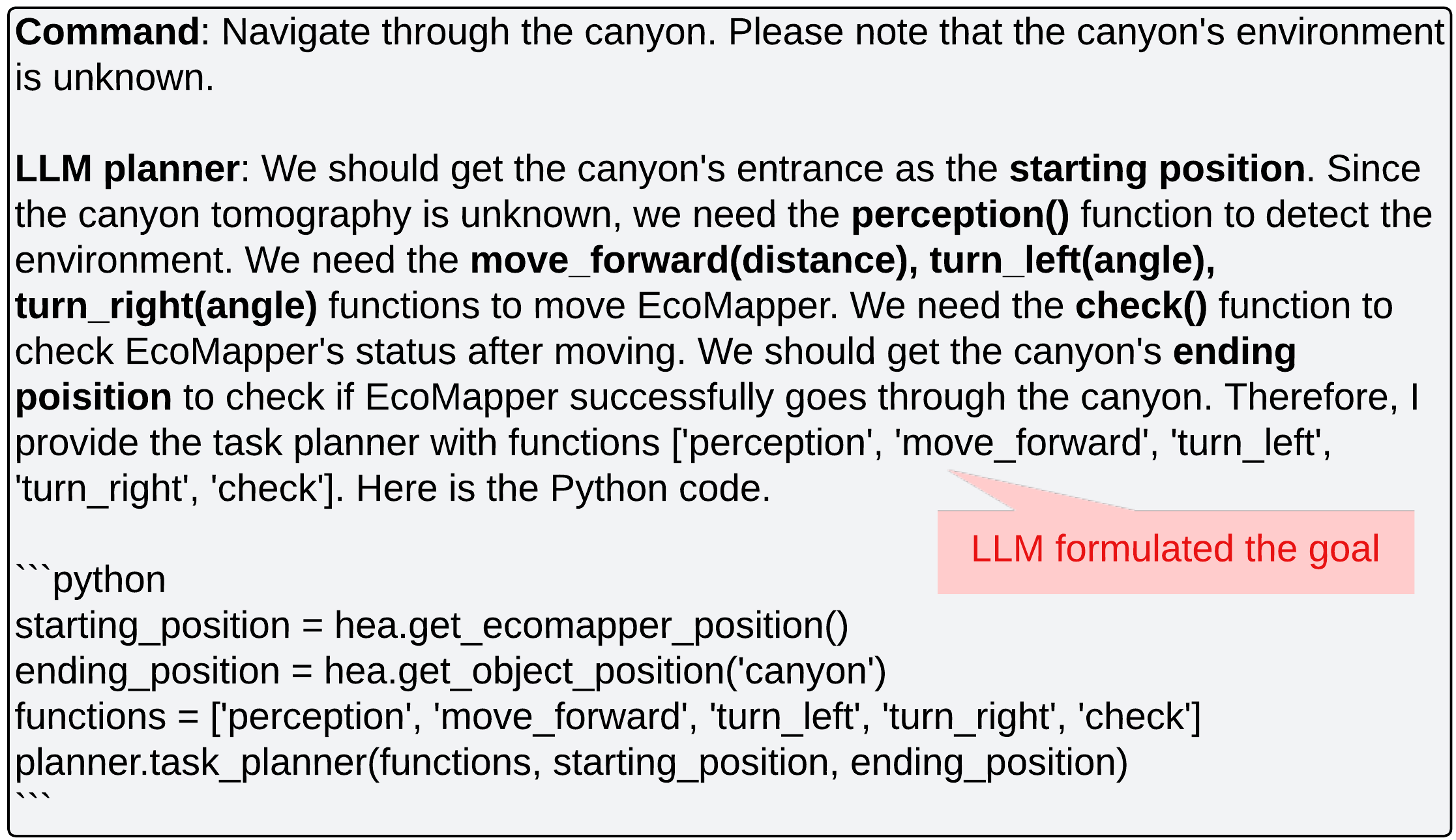}}
        \caption{Goal formulated by the LLM planner based on its semantic understating of how to finish the human command "navigating through the canyon".}
        \label{LLM generated goal}
\end{figure}

Given prompts including a set of available actions, objects, and the underwater environment, the LLM planner translates the human command into a goal to be achieved by the task and motion planners. As shown in Fig.~\ref{LLM generated goal}, the LLM planner leverages its semantic understanding to formulate the goal and select the actions for accomplishing the goal as $perception()$, $turn\_left(angle)$, $turn\_right(angle)$, $move\_forward(distance)$, and $check()$. In this work, we use GPT-4 (OpenAI) as the LLM planner.

\begin{figure}
        \centerline{\includegraphics[width=0.48\textwidth]{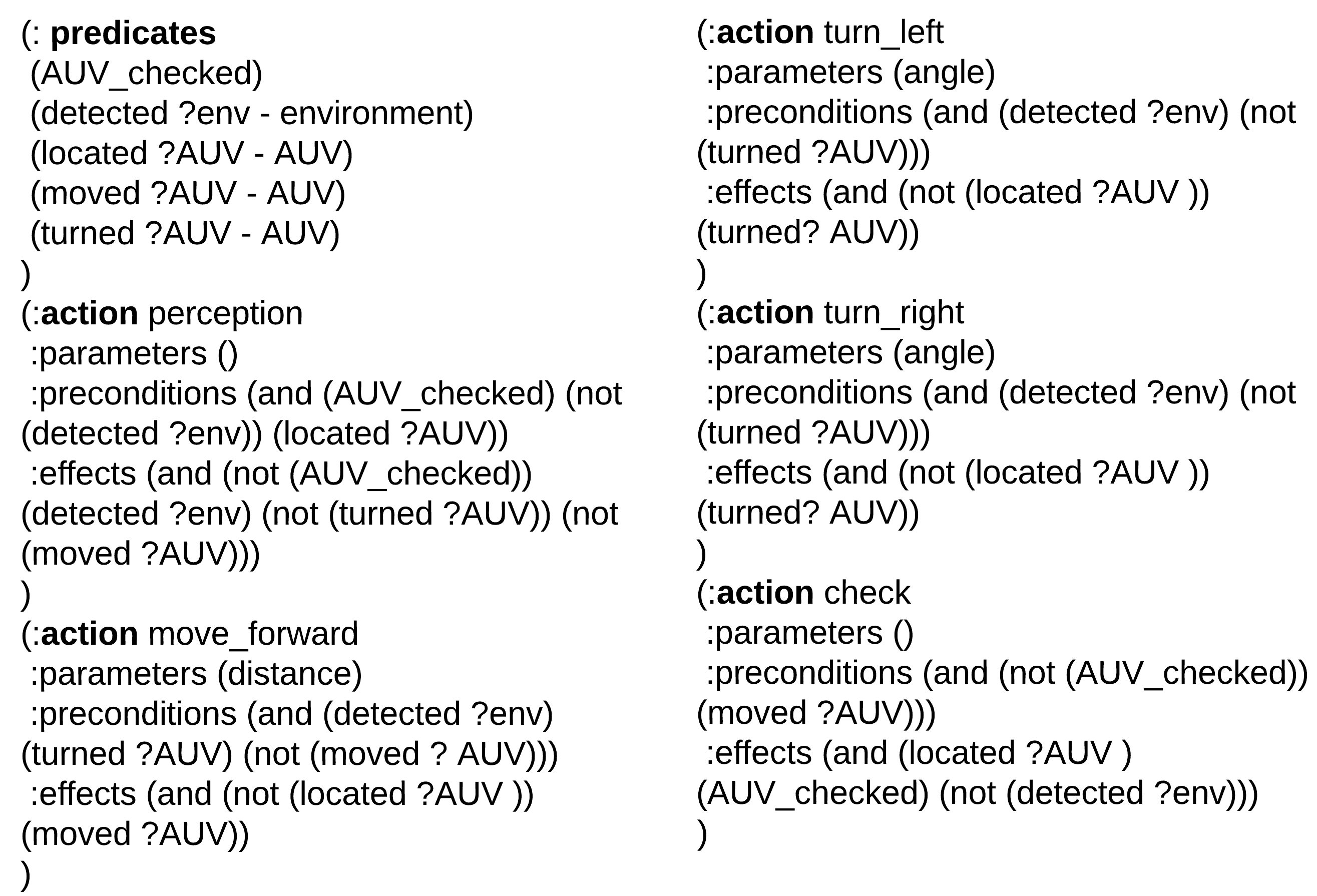}}
        \caption{Actions with predictions and effects to achieve the goal of canyon navigation.}
        \label{actions}
\end{figure}

The task planner should decide a viable task sequence out of the actions provided by the LLM planner. Given predicates representing the state space, the task planner will enhance each action with its corresponding preconditions and effects as shown in Fig.~\ref{actions}. Equipped with this logical representation, the task planner can ground the LLM-generated goal into a reasonable task sequence from the starting state to the ending state. Meanwhile, the task planner can guide the motion planner with estimated future cost represented in the depth map and laser ranges. The task planner will plan only necessary perception actions to detect the environment, so that EcoMapper moves and senses step by step to explore more regions of the unknown canyon. The observed region will only be used in the current motion planning horizon, and any part of the unobserved region is marked as an obstacle. The $check()$ action plays the role of event-triggered replanning with real-time environmental feedback. Specifically, it uses laser range sensors to check if EcoMapper is collision free. If not, the task planner regenerates a new plan for the motion planner to perform corrective actions.

The motion planner will optimize the specific motion plan of executing the task sequence, i.e., which turning angle or moving distance is best? To leverage the A* planner, we select the heuristic cost as combination of the moving distance and the distance from the rollout position to the canyon's ending position. In this way, the motion planner can perform optimal actions to make EcoMapper travel far enough distance while not deviate too much from the canyon's ending position. 

\begin{figure}
        \centerline{\includegraphics[width=0.48\textwidth]{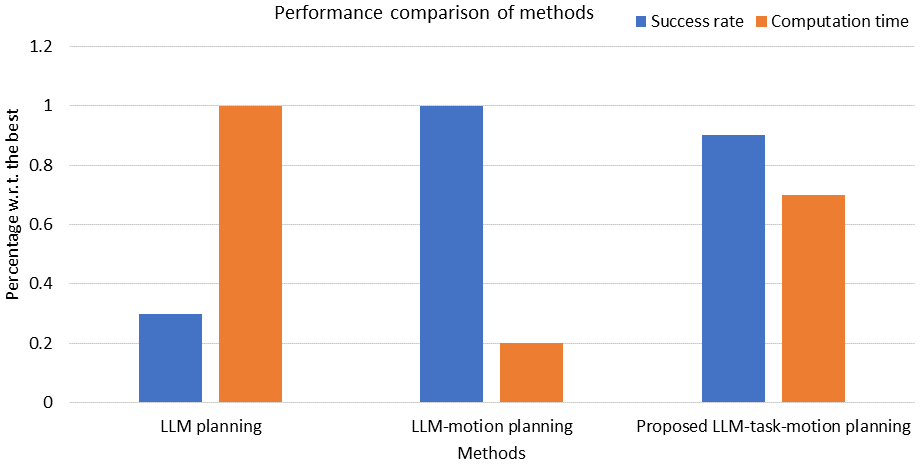}}
        \caption{Quantitative results of comparing the proposed method with two other methods. The comparison is performed in terms of both the success rate and the computation time of finishing the overall goal.}
        \label{method comparison}
\end{figure}

The LLM planner is unable to generate an executable plan because of limited real-world knowledge, while the motion planner spends excessive time delving a trajectory over long and uncertain horizons. We claim that our method can compose a valid plan and save computation time by incorporating the task planner to refine the LLM-guided goal and the motion plan according to the task logical chain and the physical dynamics, respectively. To support our claim, we compare three methods of LLM planning, LLM-motion planning and the proposed LLM-motion-task planning introduced in Section \ref{design of middle-level task planner}. The quantitative results averaging over 10 simulation runs are shown in Fig.~\ref{method comparison}. Since the LLM planning method directly composes a task sequence, it takes least computation time to achieve the goal. However, the solution quality is usually compromised, because LLMs can only consider high-level semantic representations of the tasks without any constraints. The LLM-motion planning method relies on the LLM planner to guide the motion planner for execution in the physical world. Although it can try out a solution, it takes much more computation time to exhaustively search the solution space. The proposed method accomplishes the goal with comparable success rate as the LLM-motion planning method, but with only 30\% computation time. Comparing to the LLM planning method, the proposed method takes more computation time, but it can guarantee an executable task plan.

\section{Conclusion and Future Works}
\label{conclusion}

In this paper, we present OceanChat, an AI system that utilizes LLMs' knowledge and robotic planning schemes to complete AUV missions given a human command. We incorporate three levels of LLM planning, task planning, and motion planning in a hierarchical framework. Starting with the human command, a sequential process unfolds that LLMs supply a contextually proper goal, the task planner generates a task sequence, and the motion planner seeks the optimal motion plan within the given sequence. Meanwhile an event-triggered replanning module is designed to manage unexpected execution failure. We assess the proposed system across a comprehensive AUV mission of autonomously navigating through an unknown canyon in the simulated ocean environment.

There are several promising avenues for future works. First, domain-independent PDDLStream algorithms \cite{garrett2020pddlstream, hadfield2015modular} are a powerful alternative to solve TAMP  by modeling a determinized version of  stochastic shortest path problems (SSPPs)  \cite{barry2011deth, bertsekas1996neuro, li2016act}. Second, after combining as inputs language instructions, environmental observation, and robot states, reinforcement learning \cite{sun20212d, luketina2019survey, andreas2017modular, jiang2019language, cui2022can, ma2022vip, li2023behavior, bahl2022human, ma2023liv, colas2020language, du2023guiding} can directly guide  optimization of robotic actions in an end-to-end way.  Third, our system assumes the flow effect is negligible, while zero-shot generalization \cite{oh2017zero, jang2022bc, wei2021finetuned, zeng2022socratic} can benefit from online interactions to  learn flow dynamics, effectively adapting to environmental nuances.

\newpage

\bibliography{IEEEabrv, reference.bib}
\bibliographystyle{IEEEtran}

\end{document}